\input texinfo @c -*-texinfo-*- $Id: libpa.texi,v 1.2 1997/06/10 01:18:23 ristad Exp $
@setfilename libpa.info
@settitle Library of Practical Abstractions
@set VERSION 1.2
@set RELEASE 1.2
@set ROOT libpa-1.2
@set YEAR 1997
@set UPDATED 9 June @value{YEAR}
@setchapternewpage on
@footnotestyle end
@syncodeindex vr cp
@syncodeindex fn cp
@syncodeindex tp cp

@titlepage
@sp 10
@title{Library of Practical Abstractions}
@subtitle Version @value{RELEASE}
@author Eric Sven Ristad and Peter N. Yianilos
@page
@center{Copyright Notice, License and Disclaimer}

Permission to use, copy, and distribute this software and its
documentation without fee for not-for-profit academic or research
purposes is hereby granted, provided that the above copyright notice
appears in all copies, and that both the copyright notice and this
permission notice and warranty disclaimer appear in supporting
documentation, and that neither the names of the authors nor the name of
any entity or institution to which they are related be used in
advertising or publicity pertaining to distribution of the software
without specific, written prior permission.

Permission to modify the software and documentation is hereby granted,
but only to the extent necessary to cause the software to function in a
hardware or software environment not explicitly supported as part of the
distribution, and subject to the condition that the original authors
retain all legal rights to any modifications or improvements of this
software.

The authors disclaim all warranties with regard to this software,
including all implied warranties of merchantability and fitness.  In no
event shall the authors or any entities or institutions to which he/she
is related be liable for any special, indirect or consequential damages
or any damages whatsoever resulting from loss of use, data or profits,
whether in an action of contract, negligence or other tortious action,
arising out of or in connection with the use or performance of this
software.

@vskip 0pt plus 1filll
Copyright @copyright{} @value{YEAR} Eric Sven Ristad and Peter N. Yianilos. 
@end titlepage

@ifinfo
@node Top, (dir), (dir), (dir)
@top Library of Practical Abstractions
The library of practical abstractions provides efficient implementations
of conceptually simple abstractions in the C programming language.  
This is version @value{RELEASE}, last updated @value{UPDATED}.
@cindex version

@menu
* License::             License, Copyright Notice, and Disclaimer
* Overview::            Overview
* Modules::             Library modules
* Organization::        Library organization
* Style::               Programming style
* Index::               Function and concept index
@end menu

Copyright @copyright{} @value{YEAR} by Eric Sven Ristad and Peter N. Yianilos.
All Rights Reserved.
@cindex copyright

@node License, Overview, , Top
@chapter License, Copyright Notice, and Disclaimer

Copyright @copyright{} @value{YEAR} by Eric Sven Ristad and Peter N. Yianilos.
All Rights Reserved.

Permission to use, copy, and distribute this software and its
documentation without fee for not-for-profit academic or research
purposes is hereby granted, provided that the above copyright notice
appears in all copies, and that both the copyright notice and this
permission notice and warranty disclaimer appear in supporting
documentation, and that neither the names of the authors nor the name of
any entity or institution to which they are related be used in
advertising or publicity pertaining to distribution of the software
without specific, written prior permission.  

Permission to modify the software and documentation is hereby granted,
but only to the extent necessary to cause the software to function in a
compiler or operating system environment not explicitly supported as
part of the distribution, and subject to the condition that the original
authors retain all legal rights to any modifications or improvements of
this software.

The authors disclaim all warranties with regard to this software,
including all implied warranties of merchantability and fitness.  In no
event shall the authors or any entities or institutions to which he/she
is related be liable for any special, indirect or consequential damages
or any damages whatsoever resulting from loss of use, data or profits,
whether in an action of contract, negligence or other tortious action,
arising out of or in connection with the use or performance of this
software.
@end ifinfo

@node Overview, Modules, License, Top
@chapter Overview

The library of practical abstractions (LIBPA) provides efficient
implementations of conceptually simple abstractions, in the C
programming language.  We believe that the best library code is
conceptually simple so that it will be easily understood by the
application programmer; parameterized by type so that it enjoys wide
applicability; and at least as efficient as a straightforward
special-purpose implementation.  You will find that our software
satisfies the highest standards of software design, implementation,
testing, and benchmarking.

The current LIBPA release is a source code distribution only.  It
consists of modules for portable memory management, one dimensional
arrays of arbitrary types, compact symbol tables, hash tables for
arbitrary types, a trie module for length-delimited strings over
arbitrary alphabets, single precision floating point numbers with
extended exponents, and logarithmic representations of probability
values using either fixed or floating point numbers.

We have used LIBPA to implement a wide range of statistical models for
both continuous and discrete domains.  The time and space efficiency of
LIBPA has allowed us to build larger statistical models than previously
reported, and to investigate more computationally-intensive techniques
than previously possible.  We have found LIBPA to be indispensable in
our own research, and hope that you will find it useful in yours.  If
you find LIBPA useful, please let us know!

LIBPA is not backward compatible.  We strive to provide the best
libraries for each release, and do not hesitate to completely redesign a
module to improve clarity or performance.  As a result, upgrading your
LIBPA installation to the latest release may require you to rewrite
portions of your code.  We will not change the semantics of a function
without also changing the prototype for that function.  This policy
should help you quickly upgrade your software using the latest LIBPA
release.

@menu
* Installation::        How to install and use LIBPA
* Mailing Lists::       Staying informed about LIBPA
* Acknowledgments::     Acknowledgments and affiliations
@end menu

@node Installation, Mailing Lists, , Overview
@section Installation

@menu
* Licensing::                   Licensing the source code
* Downloading::                 Getting the source code
* Environment Variables::       LIBPA environment variables
* Compiling::                   Creating object code
* Supported Environments::      Supported operating environments
@end menu

@node Licensing, Downloading, , Installation
@subsection Licensing

Every module has it's own license, written by the module's authors.
Before you install or use a LIBPA module, you must agree to the terms of
that module's license.  Typically, authors will ask that you use LIBPA
for not-for-profit research purposes only, and that you follow standard
academic practice in acknowledging your use of LIBPA in any relevant
reports or publications.

@node Downloading, Environment Variables, Licensing, Installation
@subsection Getting the Source Code

Source code for entire library may be obtained from
@example
ftp://ftp.cs.princeton.edu/pub/packages/libpa/
@end example
Please use @kbd{GNU tar -cf} to unpack the @file{.tar.gz} file.

@node Environment Variables, Compiling, Downloading, Installation
@subsection Environment Variables

Once you have ftped and unpacked the distribution, you must initialize
the following environment variables.

@table @code
@item LIBPA_GROUP
group with read/write permissions in LIBPA installation
@item LIBPA_ARCH
current machine architecture (cpu/OS combination)
@item SCC
safe compiler to produce debuggable object code
@item SLD
safe loader to produce debuggable binaries
@item UCC
unsafe compiler to produce optimized object code
@item ULD
unsafe loader to produce optimized binaries
@item RANLIB
ranlib
@item MAKE
make or gmake
@end table

The @code{LIBPA_GROUP} environment variable should be set to the name of
the group which will have read/write permissions in the installation.
We source the @file{@value{ROOT}/bin/cshrc} shell script on our system to set
these environment variables properly.@footnote{Note that we do not
recommend using gcc to compile the modules because gcc has a proprietary
implementation of the @code{assert()} macro.  Therefore, once you
compile a module using gcc, all executables based on that module must
link libgcc.a.  This is particularly annoying if you are trying to track
down a compiler bug.}

@node Compiling, Supported Environments, Environment Variables, Installation
@subsection Creating Object Code

After you have set these environment variables, run @kbd{make install}
from @file{@value{ROOT}/src} to install the libraries, and then run
@kbd{make tests} from @file{@value{ROOT}/src} in order to verify our
implementations in your operating environment.  When you are satisfied
that in the outcome of these tests, run @kbd{make clean} to delete
extraneous object code files and archives in the source directories.

As explained below, we provide safe and unsafe versions of all object
code.  The safe versions contain unoptimized object code with symbol
tables for debugging purposes.  The unsafe versions contain optimized
object code without symbol tables.  We recommend using safe object code
on a routine basis, and only using unsafe object code when performance
is of utmost importance.  The @file{libpa.a} and @file{libpa_u.a}
archives contain the safe and unsafe versions of all object code,
respectively.

@node Supported Environments, , Compiling, Installation
@subsection Supported Environments

All modules are supported on the following operating environments:

@table @asis
@item alpha
Digital Alpha with Digital UNIX 4.0
@item i586
Intel Pentium with Linux 2.0
@item sun5
Sun SPARC with SunOS 5.5 (Solaris)
@end table

Most modules support a wider range of UNIX operating environments,

including
@table @asis
@item hppa
HP PA-RISC 9000/7xx with HP-UX 9.0
@item sgi
SGI MIPS with IRIX 5.x or 6.x
@item sun4
Sun SPARC with SunOS 4.3.1
@end table

@node Mailing Lists, Acknowledgments, Installation, Overview
@section Mailing Lists

@menu 
* Announcements::       mailing list for announcements
* Help::                mailing list for help
* Bug Reports::         mailing list for bug reports
@end menu

@node Announcements, Help, , Mailing Lists
@subsection Staying Informed

@cindex announcements
We encourage you to subscribe to libpa-announce@@cs.princeton.edu, 
to learn about upgrades and new modules as they are published.  
Send mail to majordomo@@cs.princeton.edu with the single line:
@example
subscribe libpa-announce
@end example

@node Help, Bug Reports, Announcements, Mailing Lists
@subsection Getting and Giving Help

@cindex help
If you are having trouble installing or using LIBPA, or have other
questions about the library of practical abstractions, send mail to
libpa-help@@cs.princeton.edu.  If you are a skilled user of LIBPA, then
we encourage you to subscribe to this majordomo mailing list and to
help field the questions that are posted.

@node Bug Reports, , Help, Mailing Lists
@subsection Bug Reports

@cindex bug reports
We are unable to track down bugs in your code or in your operating
environment (compiler, operating system).  We have spent a significant
amount of time and care writing the LIBPA modules, and have published
the source code and test suites.  So if you think you have found a bug
in our code, please help us by sending a thoroughly tested bug fix along
with your bug report.  When you have finished testing your bug fix, we
would be grateful if you would please send it to
libpa-bugs@@cs.princeton.edu along with a detailed explanation.

@node Acknowledgments, , Mailing Lists, Overview
@section Acknowledgments

@subsection Affiliations

Eric Sven Ristad is with the Department of Computer Science, Princeton
University, 35 Olden Street, Princeton, NJ 08544.  He is partially
supported by Young Investigator Award IRI-0258517 from the National
Science Foundation.  Email: ristad@@cs.princeton.edu.

Peter N. Yianilos is with the Department of Computer Science, Princeton
University, 35 Olden Street, Princeton, NJ 08544 and with the NEC
Research Institute, 4 Independence Way, Princeton, NJ 08540.  Email:
pny@@research.nj.nec.com.

@subsection Thanks

@cindex Hanson, David
@cindex Thomas, Robert
@cindex Kanzelberger, Kirk
We are grateful to David Hanson <drh@@microsoft.com> for his feedback,
and for providing us with so many well-taught students while he was with
us at Princeton.  Robert Thomas <rgt@@cs.princeton.edu> provided
invaluable input in the design and redesign and reredesign of the LIBPA
modules.  Kirk Kanzelberger <kirk@@research.nj.nec.com> assisted with
the initial implementation of the @code{vector_t} module in Summer 1993.

@cindex Cons, Lionel
Documentation was created in GNU texinfo format, from which info and
postscript formats were derived directly. html format was derived using
texi2html from Lionel Cons.

@node Modules, Organization, Overview, Top
@chapter Library Modules

Currently we publish modules for basic memory management, tables of
arbitrary types, and extremal numerical values:

@table @code
@item memory_t
efficient portable memory management
@item vector_t
one dimensional arrays of arbitrary types	
@item table_t
compact symbol table for <key,datum> pairs.
@item trie_t
trie for mapping strings to integers
@item hash_table_t
hash table toolkit, to build your own high-performance hash tables.
@item symbol_table_t
hash table for symbols
@item string_table_t
hash table for NULL-terminated character strings
@item unigram_t
compact table of symbol frequencies with dynamic counters
@item balanced_t
single precision floating point numbers with extended exponents
@item L_t
fixed point logarithmic representation of probability values
@item logpr_t
floating point logarithmic representation of probability values
@item pr_t
generic interface to probability value libraries
@end table

@menu
* Memory::      portable memory management
* Tables::      storing arbitrary keys and values
* Numerical::   arithmetic with extreme numerical values
@end menu

@node Memory, Tables, , Modules
@section Memory Management

@cindex @code{memory_t}
The @code{memory_t} module is the foundation of our library.  All memory
management is performed in a portable manner using the @code{memory_t}
module, which helps identify memory leaks without using expensive and
nonportable third party applications, such as purify.  The
@code{memory_create()} function replaces @code{malloc()}, while the
@code{memory_destroy()} function replaces @code{free()}.  Other
functions are provided to replace the standard C memory functions.

@cindex @code{vector_t}
The @code{vector_t} module is also widely used, and is equally
indispensable.  A @code{vector_t} is a length-delimited sequence of
objects of arbitrary type.  The objects are parameterized by the size of
their representations, via the @code{sizeof()} primitive.  The
@code{vector_t} module supports dynamic resizing, concatenation,
insertion, sorting, and portable input/output.  Indeed, we often use
@code{vector_t} where LISP programmers would use lists.  Since the
@code{vector_t} objects aren't boxed and are arranged linearly in
memory, @code{vector_t} operations are at least twice as fast as lists
and require significantly less storage than lists (eg., up to 16 times
less storage for one byte objects on a machine with 8 byte pointers).

@node Tables, Numerical, Memory, Modules
@section Tables

LIBPA provides four ways to store information in tables: @code{table_t},
@code{trie_t}, @code{hash_table_t}, and @code{unigram_t}.

@cindex @code{table_t}
The @code{table_t} module is a compact table for arbitrary
<symbol,datum> pairs.  Lookups are performed in O(log n) using binary
search, while insertions and deletions are O(n) for a table containing n
entries.  The @code{table_t} module is typically used as a building
block for other modules, such as the @code{trie_t} module, where memory
usage is the single most important performance desiderata.

@cindex @code{trie_t}
The @code{trie_t} module maps length-delimited strings over arbitrary
alphabets to consecutive unsigned integers.  It also provides the
inverse map, from indices to length-delimited strings.

@cindex @code{hash_table_t}
@cindex @code{symbol_hash_t}
@cindex @code{string_hash_t}
The @code{hash_table_t} module supports the easy construction of compact
hash tables for arbitrary <key,datum> pairs with open addressing, double
hashing, and dynamic resizing.  We used the @code{hash_table_t} module
to implement two useful modules: the @code{symbol_hash_t} for arbitrary
symbols and the @code{string_hash_t} for NULL-terminated character
strings.  Again, we strove to minimize the space requirements of the
implementation so that large numbers of @code{hash_table_t} can be used
in even the most memory-intensive computations.

@cindex @code{unigram_t}
Finally, the @code{unigram_t} provides a space-efficient representation
for the frequencies of symbols drawn from an arbitrary alphabet.  Our
@code{unigram_t} implementation uses dynamic counter resizing to
minimize the amount of storage required to store the symbol frequencies.
Thus, it may be used to efficiently store the state transition
frequencies in a very large Markov model.

@node Numerical, , Tables, Modules
@section Numerical Values

@cindex @code{pr_t}
LIBPA provides three ways to represent extremely small numerical values,
such as are commonly encountered in statistical modeling applications:
@code{balanced_t}, @code{L_t}, and @code{logpr_t}.  The @code{pr_t}
module provides a generic macro interface to all three ways of
representing probability values.  The three modules offer a wide range
of performance/accuracy tradeoffs.  @code{L_t} is the fastest and
requires the least space.  @code{logpr_t} is the most accurate but the
slowest.  @code{balanced_t} is nearly as accurate as @code{logpr_t} and
can be significantly faster.

@cindex @code{balanced_t}
The @code{balanced_t} type represents single precision floating point
numbers with extended 32-bit exponents.  As a result, the
@code{balanced_t} module supports floating point arithmetic for
extremely small and extremely large numerical values -- values which
would cause overflow or underflow in double precision floating point
arithmetic.  Depending on your machine, @code{balanced_t} arithmetic is
from 10 to 20 times slower than double precision floating point
arithmetic and is only slightly less accurate that single precision
floating point arithmetic.  Unlike the other LIBPA numerical types, the
@code{balanced_t} type is capable of representing a much larger range of
numbers, such as very large numbers and negative numbers.

@cindex @code{L_t}
The @code{L_t} type represents probability values using fixed point
numbers --- @code{char}, @code{short}, @code{int}, and @code{long} ---
where larger types are capable of representing ever-smaller probability
values.  Depending on your machine, @code{L_t} arithmetic is from 1.2 to
1.8 times slower than double precision floating point arithmetic.  It is
also considerably less accurate.  @code{balanced_t} arithmetic is more
accurate than @code{L_t} arithmetic, but is more than five times slower
and typically requires twice as much memory.

@cindex @code{logpr_t}
Finally, we also provide a standard @code{logpr_t} type for the
logarithmic representation of probability values using double-precision
floating point numbers.  Although this is the most accurate
representation of probability values, it is also the slowest.  Depending
on the math libraries provided by your operating system, @code{logpr_t}
arithmetic can be 13 to 60 times slower than double precision floating
point arithmetic.

The following table reports timing results of the @file{pr.bench}
benchmark on four different machines.  All numbers are user times in
seconds, as reported by @file{/usr/bin/time} or @file{/usr/bin/timex}
(lower is better).

@cindex @code{pr_t} benchmarks
@example
                        double   logpr_t   balanced_t   L_t
                        ------   -------   ----------  -----
AlphaStation 500/500    121.8     1619.8    1292.7     207.2
Sun Ultra2 1200         301.2    10166.4    5037.2     541.9
SGI O2 180mhz           371.4     8570.8    4830.1     546.6
HP 9000/755             369.2    19643.6    7497.1     638.0
Dell GXM 5166           805.5    12010.6    5118.6     933.3
@end example
@cindex benchmarks

@node Organization, Style, Modules, Top
@chapter Library Organization

@menu
* Directories::         Directory structure
* File Names::          File naming conventions
* Test Suites::         Test suites requirements
* Benchmarks::          Benchmarks
* Object Code::         Object code and archives
* Makefiles::            required Makefile targets
@end menu

@node Directories, File Names, , Organization
@section Directory Structure

The library of practical abstractions is organized as follows:

@table @file
@item @value{ROOT}/include	
include files
@item @value{ROOT}/lib.@var{arch}
object code archives for @var{arch}
@item @value{ROOT}/src/@var{module}
source code for @var{module}
@item @value{ROOT}/bin
shell scripts
@item @value{ROOT}/bin/@var{arch}
architecture-dependent executables for @var{arch}
@item @value{ROOT}/doc
miscellaneous documentation
@end table

This allows you to compile with -I@file{@value{ROOT}/include} and link
with -L@file{@value{ROOT}/lib.@var{ARCH}}.  You may also want to put
@file{@value{ROOT}/bin} and @file{@value{ROOT}/bin/@var{ARCH}} on your
@code{PATH} environment variable.

@node File Names, Test Suites, Directories, Organization
@section File Naming Conventions

A LIBPA module named @var{module} is implemented in the following files:

@table @file
@item @var{module}.h
public prototypes and typedefs for @var{module}
@item @var{module}.c		
source code for @var{module}
@item @var{module}.test.c		
source code for @var{module} test suite
@item @var{module}.test		
safe executable for @var{module} test suite
@item @var{module}.test_u		
unsafe executable for @var{module} test suite
@item Makefile		
Makefile for @var{module}
@item license
Copyright notice, license, and disclaimer
@end table

The implementation may include other files, named according to a
similar naming convention, such as:

@table @file
@item @var{module}.bench.c
source code for @var{module} comparative benchmark
@item @var{module}.bench
executable for @var{module} comparative benchmark
@item @var{module}.pure
purified test suite for @var{module} (safe)
@item @var{module}.pure_u
purified test suite for @var{module} (unsafe) 
@end table

Special module functions may be provided a separate interface.  For
example, the @code{memory_t} library includes a @file{memory.spartan.h}
header file for the special-purpose spartan @code{memory_t} functions.
As explained below, the object code for @var{module} is placed in an
archive named lib@var{module}.a in the @file{@value{ROOT}/lib.@var{ARCH}}
directory.

@node Test Suites, Benchmarks, File Names, Organization
@section Test Suites (required)

Comprehensive test suites are provided for all code.  A test suite is
a certificate of correctness.  It must convince an aggressive skeptic
of the correctness of the implementation.  The best way to establish
correctness is to verify the behavior of the implementation against an
independent implementation.  The independent implementation should be
so simple that it is plausibly correct.  Even the weakest test suite
must exercise all functions in reasonable as well as unreasonable
operating ranges.

A test suite should run to completion in a reasonable amount of time.
It should describe the tests being performed as they are executed, but
should not display too much information either, rarely more than a
page. It should end by announcing the successful completion of the test
or by dumping core via an @code{abort()} call or an @code{assert()}
failure.

@node Benchmarks, Object Code, Test Suites, Organization
@section Benchmarks (optional)

A benchmark is a comparative performance analysis.  The goal of a
benchmark is to reveal the performance tradeoffs made in your
implementation.  Accordingly, the ideal benchmark compares the actual
time/space requirements of your implementation with a range of
alternative implementations in a realistic application.  In practice,
a benchmark should compare your implementation to at least one
alternative implementation in a realistic sequence of function calls.

@node Object Code, Makefiles, Benchmarks, Organization
@section Object Code and Archives

The library of practical abstractions provides two versions of all
object code: safe and unsafe.  We recommend the routine use of safe
object code.  Safe object code is intended for everyday use.  Safe
object code has been compiled with no optimizations, with all
@code{assert()} macros, and with full symbol tables for debugging.  Safe
libraries are named @file{lib@var{module}.a}.  The @file{libpa.a}
archive includes all safe object code in the current release.

Unsafe object code is intended for the most demanding applications.
Unsafe object code has been compiled with maximum optimizations, the the
@var{NDEBUG} macro defined, and without any symbol tables.  We strongly
discourage the routine use of unsafe object code, which is why we gave
it such a frightful name.  Under no conditions should unsafe object code
be used during debugging, because optimizations are a major source of
compiler errors and erroneous debugger information as well.  Unsafe
libraries are named @file{lib@var{module}_u.a}.  The @file{libpa_u.a}
archive includes all unsafe object code in the current release.

@node Makefiles, , Object Code, Organization
@section Makefiles

Our Makefiles include the targets "install", "tests", and "clean".  Use
@file{@value{ROOT}/src/Makefile} to perform these operations on all
modules.

@table @kbd
@item make install
Installs the header files in @file{@value{ROOT}/include} and then
installs the safe and unsafe object code archives in
@file{@value{ROOT}/lib.@var{ARCH}}.
@item make tests
Creates and runs safe and unsafe versions of the current module's test
suite.
@item make clean
Deletes all object code and archives in the current directory.
@end table

@node Style, Index, Organization, Top
@chapter Programming Style

The central goal of module implementation is correctness.  Performance
is important, but secondary to correctness.  By correctness, we mean
that all specified functions are implemented strictly according to the
module specification, without any memory leaks or segmentation faults.
Aside trivial cases, there will be no proof of correctness, only degrees
of belief in correctness.  And so the astute reader has already realized
that "correctness" really means "belief in correctness".

Our belief in the correctness of an implementation is principally
influenced by clarity of the implementation.  It is much easier to
develop confidence in code that is easy to understand than in code that
is difficult to understand.  Clarity is determined by degree of
conformance to a style guide, appropriate use of abstractions,
appropriate symbol names (that obey reasonable naming conventions),
useful but sparing comments (when essential to understanding), and
consistent formatting.  

As we hope you will see, our software places an unusual premium on
clarity, both directly and indirectly.  Our belief in the correctness of
an implementation is also affected by the rigor of the programmer's test
suite and the abundance of assertions.  Accordingly, our software always
includes a comprehensive test suite as well as aggressive use of
assertions.

@menu
* Desiderata::          Summary of coding desiderata
* Identifiers::         Identifier naming conventions
* Safety and Clarity::  Guidelines for safe and clear code
* Type Declarations::   Guidelines for type declarations
* Efficiency::          Guidelines for efficient implementations
* Memory Management::   Memory management policy
* Formatting::          Formatting standards
@end menu

@node Desiderata, Identifiers, , Style
@section Desiderata

We evaluate our code based on the three criteria of correctness,
clarity, and performance.

@itemize @bullet
@item Correctness: Does the code correctly implement the specification? 
@itemize @minus
@item all specified functions implemented strictly correctly
@item abundant use of @code{assert()} for correctness and clarity
@item a comprehensive test suite is provided by the programmer
@item no memory leaks or segmentation faults
@end itemize
@item Clarity: How easy is it to understand the code?
@itemize @minus
@item conceptually elegant specification
@item appropriate use of abstractions
@item degree of conformance to style guide
@item appropriate variable names, that obey naming conventions
@item useful comments when essential to understanding
@item proper formatting
@end itemize
@item Performance: How well does the code perform (time, space)?
@itemize @minus
@item speed of the code in real-world use
@item efficiency of memory management
@item performance relative to other implementations
@end itemize
@end itemize

@node Identifiers, Safety and Clarity, Desiderata, Style
@section Identifier Naming Conventions

Clarity is of the utmost importance in naming your identifiers.  Your
code should read like a great novel.  Replace generic names, such as
@code{cnt} or @code{count}, with maximally specific names, such as
@code{vertex_count} or @code{edge_count}.  Refrain from using
meaningless names, such as @code{temp}, @code{tmp}, or variants of the
dreaded @code{xxx}.  Single character names are reserved for integer
iteration variables.

Vowels are obligatory!  Even standard abbreviations (eg., @code{defn},
@code{dir}) should not be used unless they substantially improve
clarity.  If your typing skills are inadequate to type long names
quickly, then take a typing class, but don't burden others with your
sloth.

@menu
* Constants and Variables::     Naming conventions for constants and variables
* Addresses::                   Naming conventions for pointers and handles
* Procedures::                  Naming conventions for prototypes and types
@end menu

@node Constants and Variables, Addresses, , Identifiers
@subsection Constants and Variables

Constants created using @code{#define} should be all upper case.  Macros
created using @code{#define} should be either initial letter upper case
or all lower case (to allow replacement by procedures) when they take
arguments.  Global variables should be first character upper case.
Procedures and all local variables should be all lower case.

Compound names are formed with underscore separators, eg.,
@code{CONSTANT}, @code{nifty_macro()}, @code{Global_Variable},
@code{local_variable}.

Don't reuse variable names.  A single variable name should have a single
meaning in the widest possible scope (minimally within the body of a
procedure, preferably within an entire library).  Iteration variables
should always iterate over the same domain.

@node Addresses, Procedures, Constants and Variables, Identifiers
@subsection Pointers and Handles

When it improves clarity, variables that are pointers to a single
object should be named with the @code{_p} suffix.  Similarly, a pointer to
a pointer to an object (also called a "handle") should be named with
the @code{_pp} suffix when it improves clarity.
@example
type_t    *name_p, **name_pp;
@end example

This convention might be used, for example, to distinguish a one
dimensional array of pointers (@code{void **name_p} or @code{void
*name_p[]}) from a two dimensional array of objects (@code{void **name}
or @code{void name[][]}), or when an object is passed directly (ie., by
value) and indirectly (ie., by reference) in the same piece of code.
When an object is always passed by reference, adding the @code{_p}
suffix does not improve clarity.  Note that arrays are NOT conceptually
considered pointers, and therefore they should not be named according to
the @code{_p} or @code{_pp} suffix convention (unless the arrays are
arrays of pointers or arrays of pointers to pointers, respectively).

@node Procedures, , Addresses, Identifiers
@subsection Procedures

When possible, procedure names begin with their module name and their
prototypes obey the following naming conventions (see module design).

@menu
* Type Definitions::              Defining types
* Constructors and Destructors::  Creating and destroying objects
* Object Attributes::             Predicates, selectors, and mutators
* Type Coercion::                 Converting one type into another
* File Input/Output::             Input/output to binary and ascii files
* Argument Order::                Order of procedure arguments
@end menu

@node Type Definitions, Constructors and Destructors, , Procedures
@subsubsection Type Definitions

@table @code
@item typedef ... @var{type}_t;
typedef for objects of type @code{@var{type}_t}.
@item @var{type}_tag
internal tag for union or struct typedef.
@end table

@node Constructors and Destructors, Object Attributes, Type Definitions, Procedures
@subsubsection Constructors and Destructors

@table @code
@item @var{type}_t *@var{type}_create(...);
Creates a new object of type @code{@var{type}_t} and returns its address.
@item void @var{type}_destroy(@var{type}_t *object);
Permanently destroys an object of type @code{@var{type}_t};
all subsequent operations on that object are undefined.
@item @var{type}_t *@var{type}_copy(const @var{type}_t *object);
Creates an exact copy of an object of type @code{@var{type}_t}
and returns its address.
@item void @var{type}_initialize(@var{type}_t *address, ...);
Given the address of @code{sizeof(@var{type}_t)} bytes,
initializes that memory location to contain a valid object of type 
@code{@var{type}_t}.
@item void @var{type}_finalize(@var{type}_t *object);
Given a valid object of type @code{@var{type}_t},
finalizes that object so that the subsequent freeing
of @code{sizeof(@var{type}_t)} bytes at the given address 
will completely destroy the given object.
@end table

@node Object Attributes, Type Coercion, Constructors and Destructors, Procedures
@subsubsection Predicates, Selectors, and Mutators

@table @code
@item boolean_t @var{type}_is_@var{predicate}(const @var{type}_t *object);
Returns @code{TRUE} if and only if the Boolean @var{predicate} 
holds for a given object of type @code{@var{type}_t}.
@item value @var{type}_@var{attribute}(const @var{type}_t *object);
Returns @code{value} of given @var{attribute} for given object of type
@code{@var{type}_t}, or the address of that value.
@item void @var{type}_set_@var{attribute}(@var{type}_t *object, ...);
Change an object of type @code{@var{type}_t} so that 
the desired @var{attribute} now holds for that object.
@item void *@var{type}_workspace_p(const @var{type}_t *object);
Returns address of caller workspace in object of type
@code{@var{type}_t}, which caller is allowed to modify.
@end table

@node Type Coercion, File Input/Output, Object Attributes, Procedures
@subsubsection Type Coercion

@table @code
@item @var{typeB}_t *@var{typeA}2@var{typeB}(const @var{typeA}_t *object);
Given an object of type @code{@var{typeA}_t}, 
creates an equivalent object of type @code{@var{typeB}_t}
and returns it.
@end table

@node File Input/Output, Argument Order, Type Coercion, Procedures
@subsubsection File Input/Output

@table @code
@item boolean_t @var{type}_write(const @var{type}_t *object, FILE *stream);
Writes objects of type @code{@var{type}_t} to the binary
file stream in self-delimiting format and returns TRUE
if and only if the write succeeds.
@item @var{type}_t *@var{type}_read(FILE *stream);
Reads object of type @code{@var{type}_t} from the binary
file stream in self-delimiting format and returns it.
@item boolean_t @var{type}_fprintf(const @var{type}_t *object, FILE *stream);
Formats object of type @code{@var{type}_t} on ASCII file stream
in self-delimiting human-readable format.
@item @var{type}_t *@var{type}_fscanf(FILE *stream);
Reads object of type @code{@var{type}_t} from ASCII file stream
in self-delimiting human-readable format and returns it.
@item boolean_t @var{type}_backup(const @var{type}_t *object, FILE *stream);
Writes properties of an object of type @code{@var{type}_t} to the binary
file stream in self-delimiting format and returns TRUE if and only if
the write succeeds.  This is used when
@item boolean_t @var{type}_restore(@var{type}_t *object, FILE *stream);
Reads properties of an object of type @code{@var{type}_t} from the
binary file stream in self-delimiting format and returns TRUE if and
only if the read succeeds and the objects properties are updated from
the file stream.
@end table

@node Argument Order, , File Input/Output, Procedures
@subsubsection Argument Order

The first argument to an function should always be an object of the
type defined in the function's module, eg., 
@example
@code{void   @var{type}_write(const @var{type}_t *, FILE *);}
@code{@var{type}_t *@var{type}_read(FILE *);}
@end example

@node Safety and Clarity, Type Declarations, Identifiers, Style
@section Safety and Clarity

Avoid macros!  Macros are a major source of bugs.  Use procedures
instead.  Once your code is working, and you find yourself with extra
time on your hands, then you can replace some procedures with macros for
minor performance gains, although we still do not recommend wasting your
time this way.

All source code should make abundant use of the @code{assert()} macro
from the @file{<assert.h>} library for both clarity and correctness.
Correctness includes checking that procedures are passed the proper
arguments and that they return proper values.  Clarity includes
asserting all known invariants and assumptions.  As a rough guide,
around 50
performing pointer arithmetic.  No kidding.  The source code must work
properly even if @code{NDEBUG} is defined and all the @code{assert()}
macros vanish.  That means no side-effects may occur in the expressions
passed to @code{assert()}.

Replace compound asserts with primitive asserts in order to make
debugging easier.  For example, the compound assertion
@example
@code{assert(exp1 && exp2);}
@end example
should be replaced with the two primitive assertions
@example
@code{assert(exp1); assert(exp2);}
@end example

Do not mutate a variable inside a complex expression or when it appears
as an argument to a procedure call.  If this restriction makes you whine
about typing effort, it's time to take another typing class.  Thus,
@code{f(i++);} should be @code{f(i); i++}.

All blocks must be explicitly delimited by curly brackets, especially if
the blocks contain exactly one statement, eg.,
@example
@code{if (predicate) @{ statement; @}}
@end example

Liberal use of parentheses is encouraged, particularly within macro
definitions.  Programming is not a test to see whether you know all
the most compact legal expressions of C.  Rather, it is a test to see
if you can write correct code of breathtaking clarity.

Cutting and pasting code is likely to introduce errors.  It is also an
early warning sign of a bad modularization.  So don't do it!  If you are
unable to break your cut-and-paste habit, then you must review newly
pasted code with extraordinary diligence.

@node Type Declarations, Efficiency, Safety and Clarity, Style
@section Type Declarations

Whenever possible, arguments passed by reference (ie., pointers)
should also be declared const.  Not only does this improve the clarity
and safety of your code, but it allows other programmers to use your
code.  A procedure argument that is not declared const can never be
const in any code that uses that procedure.

Do not mix declarations and initializations.  That is, do not combine
the type declaration of automatic variables with their initialization in
the body of a procedure.  For example, replace @code{int i=0;} with the
declaration @code{int i;} in the preamble followed by the actual
initialization @code{i=0;} in the body of the procedure.

Declare variables in the narrowest possible scope.  If you only use a
variable inside one of the branches of a conditional statement, then
it should be declared only in that branch.  This reduces the risk of
improperly initialized variables and inadvertent name conflicts.  It
also makes the code easier to understand.

In general, typedefs should be to the object itself, rather than to a
pointer to the object.  This makes memory management easier in most
cases, and improves clarity because the caller better understands the
semantics of procedure arguments.

Objects requiring more space than two machine words should typically be
passed by reference for the purposes of efficiency.  However, objects
that are used solely to return multiple values from a procedure call
should typically be returned by value in order to simplify memory
management.

We strongly encourage you to design modules that are parameterized by
types.  Type parameterization allows you to design efficient, reusable
modules.  In C, types are parameterized by the size of their objects (in
bytes) along with a minimal set of procedures necessary to operate on
those objects.

@section Additional Guidelines
@node Efficiency, Memory Management, Type Declarations, Style
@subsection Efficiency

No operating system calls, such as memory allocation or file i/o, should
occur in time-critical inner loops.  This may require design changes.

Minimize the number of pointers in your data structures.  Each pointer
reference requires a potentially-nonlocal memory reference, resulting in
a cache miss, and can take up to 8 bytes of storage on some machines.

@node Memory Management, Formatting, Efficiency, Style
@subsection Memory Management

Even a small memory leak is unacceptable.

All LIBPA code must use the @code{memory_create()} and
@code{memory_destroy()} procedures from the @code{memory_t} library
instead of the @code{malloc()} and @code{free()} provided by the
standard @code{stdlib}.

Every test suite must conclude with a memory leak check.
@example
@code{assert(memory_total_bytes() == 0);}
@code{assert(memory_total_blocks() == 0);}
@end example

@node Formatting, , Memory Management, Style
@subsection Formatting
	
Use GNU emacs C mode.  In most cases, curly braces should be on their
own line, in order to more clearly delineate the scope of the block.

@node Index, , Style, Top
@unnumbered Index
@printindex cp

@contents
@bye